\documentclass{article}

\usepackage{microtype}
\usepackage{graphicx}
\usepackage{subcaption}
\usepackage{booktabs} 
\usepackage{hyperref}

\usepackage[preprint]{icml2026}

\definecolor{oursbg}{HTML}{E0F6FF} 
\definecolor{graytext}{HTML}{808080} 
\definecolor{headergray}{HTML}{F0F0F0} 
\definecolor{evengray}{HTML}{FAFAFA} 
\definecolor{inc-red}{HTML}{E05E5E} 
\definecolor{dec-blue}{HTML}{3490DE}

\newcommand{\up}[1]{\textcolor{inc-red}{\scriptsize\ensuremath{\uparrow}#1}}
\newcommand{\down}[1]{\textcolor{dec-blue}{\scriptsize\ensuremath{\downarrow}#1}}
\newcommand{\normalup}[1]{\textcolor{inc-red}{\normalsize\ensuremath{\uparrow}#1}}
\newcommand{\normaldown}[1]{\textcolor{dec-blue}{\normalsize\ensuremath{\downarrow}#1}}

\usepackage{xspace}
\makeatletter
\DeclareRobustCommand\onedot{\futurelet\@let@token\@onedot}
\def\@onedot{\ifx\@let@token.\else.\null\fi\xspace}
\def\eg{\emph{e.g}\onedot} 
\def\ie{\emph{i.e}\onedot} 
\def\etc{\emph{etc}\onedot} \def\vs{\emph{vs}\onedot}

\makeatother

\usepackage{amsmath}
\usepackage{amssymb}
\usepackage{mathtools}
\usepackage{amsthm}

\usepackage[capitalize,noabbrev]{cleveref}

\theoremstyle{plain}

\theoremstyle{definition}

\theoremstyle{remark}

\usepackage{tabularray} 
\usepackage{tcolorbox}
\usepackage{fancyvrb} 
\usepackage{fvextra} 
\usepackage{bbm}
\usepackage{float} 
\usepackage{listing}
\usepackage[dvipsnames]{xcolor}

\usepackage[textsize=tiny]{todonotes}

\icmltitlerunning{Active Exploring like a Pigeon: Reinforcing Spatial Reasoning via Agentic Vision-Language Models}

\begin{document}

\twocolumn[
  \icmltitle{Active Exploring like a Pigeon: \\Reinforcing Spatial Reasoning via Agentic Vision-Language Models}

  \icmlsetsymbol{equal}{*}

  \begin{icmlauthorlist}
    \icmlauthor{Wei Deng}{network,bupt}
    \icmlauthor{Xianlin Zhang}{bupt,art}
    \icmlauthor{Mengshi Qi*}{network,bupt}
  \end{icmlauthorlist}

  \icmlaffiliation{network}{State Key Laboratory of Networking and Switching Technology}
  \icmlaffiliation{art}{School of Digital Media \& Design Arts}
  \icmlaffiliation{bupt}{Beijing University of Posts and Telecommunications, China}

  \icmlcorrespondingauthor{Mengshi Qi}{qms@bupt.edu.cn}

  \icmlkeywords{VLM, Spatial Reasoning, Reinforcement Learning}

  \vskip 0.3in
]
\printAffiliationsAndNotice{}  

\begin{abstract}
Enabling Vision-Language Models (VLMs) to perform spatial reasoning remains challenging.
Existing approaches treat VLMs as passive observers, which is difficult for real-world applications.
Moreover, reinforcement learning methods rely on sparse rewards, limiting their effectiveness for complex reasoning tasks.
Inspired by pigeons' building and exploiting cognitive maps for navigation, we propose a novel agentic pipeline for spatial reasoning.
First, we introduce a new \emph{dynamic cognitive map} parameterizing scene layout as object positions and orientations, serving as persistent memory for new observations.
Second, we propose a novel \emph{Spatial Assertion Codes (SAC)}, Python expressions programmatically describing spatial relationships.
By collaborating with the dynamic cognitive map, SAC enables verification of intermediate reasoning steps, providing dense reward signals.
We optimize the model via supervised and reinforcement finetuning.
Experiments on the MindCube benchmark demonstrate state-of-the-art performance with \emph{80.5\%} overall accuracy, outperforming the best current method by \emph{29.5} accuracy points (a relative improvement of \emph{53.2\%}) on the challenging \textsc{Rotation} subset. Our code and data are open-sourced at \url{https://github.com/dw-dengwei/active-spatial-reasoning.git}.
\end{abstract}
\section{Introduction}
\begin{figure}
    \centering
    \includegraphics[width=.9\linewidth]{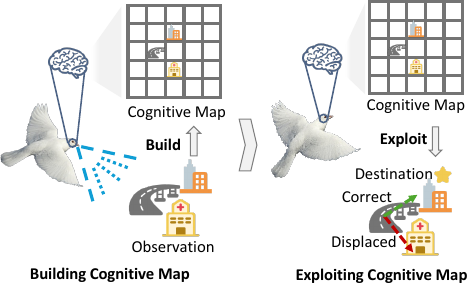}
\caption{ Illustration of the active exploring like a pigeon.
\textbf{(Left)} The pigeon can build a cognitive map from observations in mind.
\textbf{(Right)} The cognitive map guides the pigeon to navigation.
}
\label{fig:teaser}
\end{figure}

Large Language Models (LLMs) have demonstrated remarkable capabilities in various linguistic tasks, such as problem-solving~\cite{cot,tot}, code generation~\cite{code_gen}, \etc.
Extending this intelligence to the visual domain, Large Vision-Language Models (VLMs)~\cite{qwen2.5-vl,GPT-4o,InterVL} have shown impressive performance in visual understanding and reasoning~\cite{mcot,Skywork,r1-onevision}.
Despite these advancements, enabling VLMs to perform spatial reasoning remains a significant challenge.
Spatial reasoning involves perceiving spatial relationships among objects and understanding their dynamic evolution from visual observations~\cite{vsibench}, which plays a crucial role in real-world applications, such as embodied AI and robotics~\cite{EmbodiedScan,GibsonEnv,Open_X_Embodiment}.
For example, given some first-person views that observed by an embodied agent, determining what is to the left of the agent after a series of movements and rotations starting from an initial view.
The agent must integrate visual cues from the separate views and simulate these movement and rotation actions to answer the question.

Existing approaches~\cite{SpatialMLLM,mindcube,3DThinker} typically treat VLMs as passive observers, providing them with all available visual information as context and then prompting VLMs to reason over the entire scene.
This paradigm of passively perceiving the whole environment is inefficient and impractical for real-world applications, where scenes can be large and contain substantial task-irrelevant information.
To advance embodied agents from passively reasoning on images to operating effectively in real-world scenarios, it is essential to model VLMs as active explorers capable of selectively perceiving the environment according to task demands, a paradigm known as active perception.

On the other hand, reinforcement learning is widely used in the training of LLMs and VLMs. Recent advances in reinforcement learning with verifiable rewards (RLVR)~\cite{GRPO} have shown promise in optimizing VLMs for spatial reasoning tasks~\cite{VILASR,Thinking-in-360,CoV}.
These approaches successfully reinforce VLMs in spatial reasoning via Grouped Reward Policy Optimization (GRPO)~\cite{GRPO}, which rolls out multiple responses for each input and assign rewards to them to guide the optimization.
However, their reinforcement finetuning processes rely on the final correctness, providing sparse feedback, leading to suboptimal performance. Constructing dense feedback is particularly difficult because natural language outputs is flexible but unstructured, making it hard to evaluate their correctness of natural language corresponding to an underlying visual scene. Despite impressive progress on LLM-as-a-judge, which uses LLMs to evaluate responses and provide dense feedback signal, LLMs have inherent weaknesses such as hallucination and overconfidence.
The issues undermine the reliability when used for visual spatial reasoning~\cite{llm-as-a-judge}.
Consequently, it is essential to develop a computable dense-reward mechanism in spatial reasoning.

In this work, as shown in \cref{fig:teaser}, drawing inspiration from the biological evidence that homing pigeons can construct cognitive maps that include man-made structures like highways as landmarks, and correct the route when displaced by exploiting the map-like spatial memory representation~\cite{Lipp2004,bingman2006behavioral}, we propose to encode the visual observations as structured cognitive maps and translate the natural language spatial reasoning into executable Python code. 
Specifically, we propose a new dynamic cognitive map that parameterized the spatial layout of a scene as positions and orientations of the objects.
The cognitive map serves as a persistent memory that continually integrates new observations during the agentic perception process.
This process is much like a pigeon exploring a scene and building a mental map based on landmarks.
Furthermore, we introduce a novel Spatial Assertion Code (SAC), a Python-based expression that programmatically describes the spatial relationships among objects.
It mirrors the pigeon's refinement of navigation strategies through mental experience.
We collaboratively leverage the parameterized cognitive map and SAC to verify if the spatial reasoning steps are correct.
To this end, we achieve the objective of dense reward signals to reinforce the VLM for spatial reasoning.
In summary, our contributions are three-fold:

(1) We propose a novel agentic spatial reasoning pipeline within VLMs that maintains a dynamic cognitive map, a updatable memory parameterizing the spatial layout.

(2) We introduce the new Spatial Assertion Code (SAC) that collaborates with the dynamic cognitive map to verify the correctness of intermediate spatial reasoning, providing dense reward signals for reinforcement learning.

(3) Our model achieves state-of-the-art performance on the MindCube dataset~\cite{mindcube}, surpassing the best existing work by a relative improvement of 7.0\%, and by \emph{29.5} accuracy score (a relative improvement of \emph{53.2\%}) on the challenging \textsc{rotation} subset.
\section{Related Work}
\noindent\textbf{Spatial Cognition in VLMs.}
Integrating spatial intelligence into VLMs has become an important research direction, aiming to enable AI systems to understand and reason in space.
The primary research direction is to directly enhance existing VLMs so that they have the capability to process and understand spatial information.
For example, LLaVA-3D~\cite{LLaVA-3D} proposes a `3D Patch' representation integrating CLIP features with spatial coordinates.
Scene-LLM~\cite{SceneLLM} employs a hybrid 3D visual representation integrating scene-level and egocentric information.
Spatial-MLLM~\cite{SpatialMLLM} introduces a spatial encoder initialized with a 3D foundational model (VGGT~\cite{VGGT}) to capture 3D structural features, enabling visual-based spatial reasoning from purely 2D observations.
Recently, MindCube~\cite{mindcube} proposes a static cognitive map for spatial reasoning.
3DThinker~\cite{3DThinker} leverages a 3D reconstruction auxiliary task under a passive perception pipeline.
HATCH~\cite{HATCH} learns transformation relationships between pairs of views for spatial reasoning.
In contrast, we propose an agentic pipeline that actively integrates multi-view observations into a dynamic cognitive map.

\begin{figure*}[t]
    \centering
    \includegraphics[width=\textwidth]{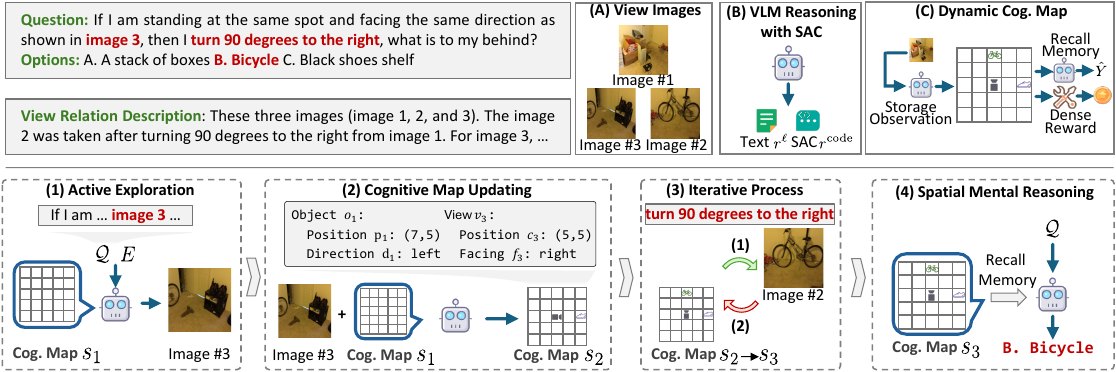}
    \caption{
    \textbf{(Top)} Question $\mathcal{Q}$, view transformation relationships $E$ are given.
    (A) There are multiple view images $V=\{v_n\}_{n=1}^{N}$ provided to be perceived by the VLM.
    (B) Our VLM outputs SAC alongside the natural language reasoning when performing spatial reasoning.
    (C) We propose the dynamic cognitive map that stores observations, recalls memory by the VLM, and computes dense rewards collaborating with SAC.
    \textbf{(Bottom)} Our model (1) actively explore the scene to retrieve related views and (2) updates a continually updated cognitive map.
    (3) The two steps are repeated iteratively.
    (4) Finally, it reasons about the answer $Y$ based on the cognitive map.
    }
    \label{fig:main_figure}
\end{figure*}

\noindent\textbf{Reinforcement Learning with Verifiable Reward.}
Reinforcement Learning (RL) has been successfully used to enhance the reasoning capabilities of Large Language Models (LLMs) and Vision-Language Models (VLMs)~\cite{InstructGPT,PPO,DPO,GRPO}.
A notable example is the Group Relative Policy Optimization (GRPO) algorithm, applied in DeepSeek-R1~\cite{GRPO}, which estimates a baseline through relative rewards within groups.
Building on this innovation, recent studies have extended GRPO-based methods to spatial reasoning tasks. 
Reason-RFT~\cite{Reason-RFT} applies GRPO-based RL to generate diverse reasoning-response pairs, significantly boosting the model's robustness against domain shift.
Similarly, SpaceR~\cite{SpaceR} introduces a method called ``Spatially-Guided Verifiable Reward Reinforcement Learning'', which also builds on the GRPO framework. However, the reward design in these works primarily focuses on final accuracy, neglecting the unique challenges of spatial reasoning tasks that require dense rewards. In this work, we propose to leverage code generation to provide dense progress rewards.

\section{Preliminary}

\textbf{Problem Formulation.} Given a question $\mathcal{Q}$, a set of limited views $V=\{v_n\}_{n=1}^{N}$, and the corresponding view transformation relationships $E=\{e_{n,m}| n,m\in[1,N]\}$, where $e_{n,m}$ describes how to transform view $v_n$ to view $v_m$ via free-form text descriptions, we aim to train a VLM to reason about the answer $Y$ corresponding to the question $\mathcal{Q}$ and the view images $V$. 

We formulate the task as a sequential decision-making problem defined by the tuple $(\mathcal{S}, \mathcal{A}, \mathcal{P}, \mathcal{R})$.
Here, $\mathcal{S}$ denotes the state space represented by a cognitive map, a continually updated memory of the scene.
The cognitive map is dynamically refined throughout the process by the state transition function.
$\mathcal{A}$ is the action space.
An action is sampled from a policy model $\pi_\theta(a_{t+1}|s_t, \mathcal{Q}, E)$, which is parameterized by a VLM.
$\mathcal{P}$ is the state transition function, which updates the state based on an action and an observation: $s_{t+1} = \mathcal{P}_\theta(s_t,v_{t+1},a_{t+1})$.
$\mathcal{R}$ is the reward function that evaluates the quality of the trajectory $R=R(a_T, s_T, \cdots, a_1, s_1)$.
The final action $a_T$ produces the answer $\hat{Y}$.
Our objective is to optimize the parameter $\theta$ of the policy model $\pi_\theta$ that maximizes the expected cumulative reward.

\textbf{Dynamic Cognitive Map.}
\cref{fig:main_figure} (C and 2) illustrate our proposed dynamic cognitive map.
The dynamic cognitive map $s_t$ is a parameterized representation of the scene spatial layout maintained within a unified top-down-view coordinate system.
It can be used to store the observations, recall memory for reasoning and compute dense reward with its layout parameters.
Formally,
\begin{equation}
s_t = \{\mathcal{O}_t, \mathcal{V}_t\},
\end{equation}
where $\mathcal{O}_t = \{(o_i, \mathbf{p}_i, \mathbf{d}_i)\}_{i=1}^{|\mathcal{O}_t|}$ denotes the set of objects with their positions $\mathbf{p}_i \in \mathbb{R}^2$ and orientations $\mathbf{d}_i$,
 and $\mathcal{V}_t = \{(v_j, \mathbf{c}_j, \mathbf{f}_j)\}_{j=1}^{|\mathcal{V}_t|}$ represents camera viewpoints with positions $\mathbf{c}_j$ and facing directions $\mathbf{f}_j$.

\textbf{Spatial Assertion Code (SAC).}
Since it is difficult to directly evaluate the correctness of natural language reasoning, we propose the concept of Spatial Assertion Code (SAC) that translates the natural language reasoning into executable code.
For example, the SAC \textit{obj1 in obj0.left(view=v4)} is a Python expression that is translated from the natural language reasoning \texttt{From view 4, object 1 is to the left of the object 0}.
Formally, let $r^\ell$ be a natural language reasoning text that describes spatial relationships among objects. We term the VLM-generated executable code $r^\textrm{code}$ as SAC, where $r^\textrm{code}$ is sampled from the model's probability distribution $p$ over code outputs conditioned on $r^\ell$:
\begin{equation}
r^\textrm{code} \sim p(r^\ell),
\end{equation}
where $r^\textrm{code}$ is a Python expression with a boolean output (\texttt{True} or \texttt{False}).

\begin{figure*}[t]
    \centering
    \includegraphics[width=0.9\textwidth]{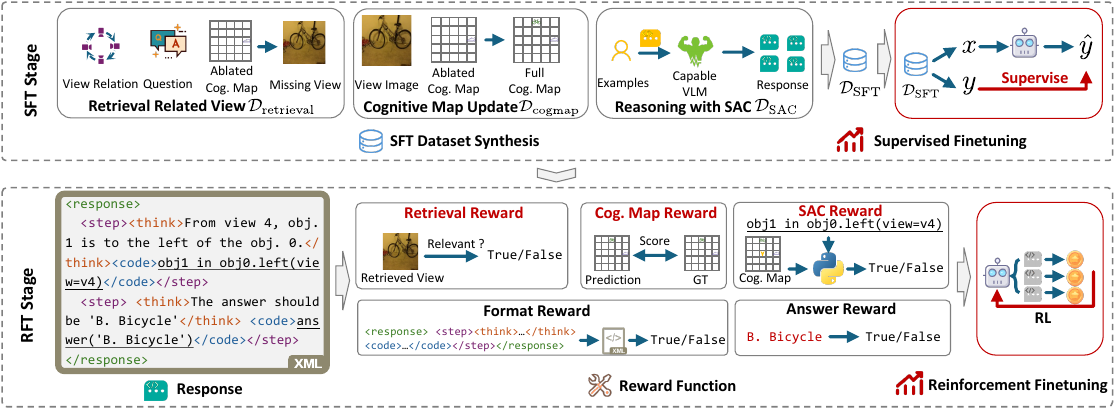}
    \caption{
    Two-stage training process of our model.
    \textbf{(Top) Supervised Finetuning.}
    We synthesize dataset with aspects of related view retrieval, dynamic cognitive map updating, and spatial reasoning with SAC for supervised finetuning.
    \textbf{(Bottom) Reinforcement Finetuning.}
    We define a reward function that measures the retrieval relatedness, cognitive map correctness, and spatial reasoning correctness for reinforcement finetuning.
    }
    \label{fig:train}
\end{figure*}

\section{Methodology}
In this section, we first present our agentic spatial reasoning pipeline in \cref{sec:pipeline}, followed by an explanation of the training process in \cref{sec:training}.
\subsection{Agentic Spatial Reasoning Pipeline}\label{sec:pipeline}

As illustrated in \cref{fig:main_figure}, our agentic spatial reasoning pipeline begins by (1) actively exploring the scene, (2) updating the cognitive map, (3) iteratively repeating the above two steps, and (4) finally performing spatial mental reasoning to answer the question.

\noindent\textbf{(1) Active Exploration.}
We use the policy model, parameterized by $\theta$, to reason about the view index based on the current state $s_t$, question $\mathcal{Q}$, and the view transformation relationships $E$:
\begin{equation}
a_{t+1} \sim \pi_\theta(s_t, \mathcal{Q}, E),
\end{equation}
where $s_t$ is the current cognitive map.
The retrieval action $a_{t+1}$ is expressed as Python code.
For example, \texttt{retrieve(3)} means retrieving the third view in $V$.

\noindent\textbf{(2) Cognitive Map Updating.}
Afterwards, we use the same VLM to update the current cognitive map $s_t$ to a new state $s_{t+1}$ based on the observed view image $v_{t+1}$:
\begin{equation}
s_{t+1} = \mathcal{P}_\theta(s_t, v_{t+1}, a_{t+1}),
\end{equation}
where $s_t$ is the current cognitive map, and $a_{t+1}$ is the instruction that prompts the VLM to integrate observation $v_{t+1}$ when updating $s_t$.

\noindent\textbf{(3) Iterative Process.}
The above two steps are repeated until the model decides to stop retrieving more views and answer the question $\mathcal{Q}$ or the maximum number of retrieval steps $T_{\textrm{max}}$ is reached.

\noindent\textbf{(4) Spatial Mental Reasoning.}
After gathering information, our framework employs the policy model $\pi_\theta$ to answer the question $\mathcal{Q}$.
The VLM recalls the accumulated dynamic cognitive map $s_t$ and generates the answer $\hat{Y}$.

Note that we innovatively introduce the dynamic cognitive map $s_t$ as a continually updated memory in the agentic pipeline.
Besides, we propose to enforce the VLM to generate SAC $r^{\textrm{code}}$ alongside natural language reasoning $r^\ell$ during the whole reasoning process.

\subsection{Training Process}
\label{sec:training}

In this section, we elaborate on the training process of our model, as illustrated in \cref{fig:train}.
It is difficult to emerge the capability of agentic spatial reasoning with SAC from a base VLM solely relying on prompt engineering without any finetuning.
Consequently, we design a two-stage post-training process that combines supervised fine-tuning (SFT) and reinforcement learning finetuning (RFT), to endow the VLM with the ability of agentic spatial reasoning with SAC.

\noindent\textbf{Supervised Finetuning.}
We perform SFT for the base VLM on a cold-start training dataset $\mathcal{D}_{\mathrm{SFT}}$ to initialize the capability of agentic spatial reasoning with SAC.
The dataset $\mathcal{D}_{\mathrm{SFT}}$ contains three types of training pairs: (1) related view retrieval; (2) cognitive map updating; and (3) spatial reasoning with SAC, mirroring the steps in our agentic spatial reasoning pipeline in \cref{fig:main_figure}:
\begin{equation}
\mathcal{D}_{\mathrm{SFT}} = \mathcal{D}_{\mathrm{retrieval}} \cup \mathcal{D}_{\mathrm{cogmap}} \cup \mathcal{D}_{\mathrm{SAC}},
\end{equation}
where $\mathcal{D}_{\mathrm{retrieval}}$ is constructed by ablating the information of a relevant view in the ground truth cognitive map, supervising the model to retrieve the missing view based on the question and the ablated cognitive map.
$\mathcal{D}_{\mathrm{cogmap}}$ is built by providing a cognitive map with partial information of a scene and a view image, supervising the VLM to integrate the partial cognitive map and the image to produce a more complete cognitive map.
$\mathcal{D}_{\mathrm{SAC}}$ is created by prompting a powerful proprietary VLM to reason with SAC, using several manually annotated examples as in-context demonstrations.

Afterwards, we perform SFT on the policy model $\pi_\theta$ with the following loss function:
\begin{equation}
\mathcal{L}_{\mathrm{SFT}} = -\mathbb{E}_{(x, y) \sim \mathcal{D}_{\text{SFT}}} \left[ \log p_\theta(y | x) \right].
\end{equation}
As a result, the SFT stage optimizes the base VLM parameterized by $\theta$ to a model capable of agentic spatial reasoning with SAC, termed the SFT model parameterized by $\theta_{\mathrm{SFT}}$.

\noindent\textbf{Reinforcement Finetuning.}
After cold-start training, we finetune the SFT model $\pi_{\theta_{\mathrm{SFT}}}$ with RL.
The RL stage relies on extensive rollouts of the SFT model, followed by a reward function that directs the RL algorithm to reinforce spatial reasoning.
We first introduce our reward function, followed by the RL algorithm.

In order to provide dense feedback, we design a new reward function that evaluates the response from three primary aspects: 1) retrieval relatedness; 2) cognitive map correctness; and 3) spatial reasoning correctness.
Formally, the reward function is defined as follows:
\begin{equation}
R = \mathbbm{1}_{\textrm{correct}}\cdot \\
\left[\mathbbm{1}_{\textrm{correct}}+w\cdot(R_{\textrm{retrieval}}+R_{\textrm{cogmap}}+R_{\textrm{SAC}})\right],\label{eq:reward_function}
\end{equation}
where $\mathbbm{1}_{\textrm{correct}}$ equals 1 if the final answer is correct, and 0 otherwise.
If the final answer is incorrect, the total reward is 0, preventing reward hacking where the algorithm optimizes to reach high reward on other aspects regardless of the final task accuracy. $R_{\textrm{retrieval}}$ scores if the retrieved views are related to the question based on the ground-truth $Y$ and the meta information from the dataset.
$R_{\textrm{cogmap}}$ evaluates if the cognitive map is correctly compared with the ground truth cognitive map.
$R_{\textrm{SAC}}$ assesses if the intermediate spatial reasoning is correct via code generation:
\begin{equation}
R_{\textrm{SAC}} = \frac{1}{M}\sum_{i=1}^{M} \mathbbm{1}({\textrm{eval}(\textrm{code}_i, s_t)==\texttt{True}}),
\end{equation}
where $\textrm{eval}(\textrm{code}_i, s_t)$ runs the $\textrm{code}_i$ given the cognitive map context $s_t$. $M$ is the number of intermediate spatial reasoning steps.

Based on the reward function, we optimize the policy model $\pi_\theta$ using the GRPO algorithm~\cite{GRPO}:
\begin{equation}
\begin{aligned}
\mathcal{L}_{\text{RL}} = \mathbb{E}_{\substack{\{\tau_i\}_{i=1}^{G}\sim\pi_{\theta_{\text{old}}}}}
\Bigg[
-\frac{1}{G}\sum_{i=1}^{G}\frac{1}{T_i}\sum_{t=1}^{T_i}
\min\Big(
\rho_{i,t}A_i,\; \\
\qquad \operatorname{clip}(\rho_{i,t},\,1-\epsilon,\,1+\epsilon)A_i
\Big)
\Bigg],
\end{aligned}
\label{eq:grpo_objective}
\end{equation}
where each rollout trajectory is $\tau_i=\{(a_{i,t}, s_{i,t})\}_{t=1}^{T_i}$ sampled from the frozen policy $\pi_{\theta_{\text{old}}}$, and
\begin{equation}
\begin{aligned}
\rho_{i,t}
&= \frac{\pi_{\theta}\big(a_{i,t} \mid s_{i,t-1}, \mathcal{Q}, E\big)}
{\pi_{\theta_{\text{old}}}\big(a_{i,t} \mid s_{i,t-1}, \mathcal{Q}, E\big)},
\\
A_i
&= \frac{S_i - \operatorname{mean}(\{S_j\}_{j=1}^{G})}
{\operatorname{std}(\{S_j\}_{j=1}^{G})}.
\end{aligned}
\label{eq:grpo_rho_adv}
\end{equation}
Here $S_i = R(\tau_i)$ is the reward of trajectory $\tau_i$ computed with \cref{eq:reward_function}, $G$ denotes the group size, $T_i$ is the valid reasoning length of $\tau_i$, and $\epsilon$ is the clipping coefficient. The ratio $\rho_{i,t}$ compares the current policy with the frozen one, while the normalized advantage $A_i$ highlights relative high-quality trajectories within the group.
We denote the model after RFT as $\pi_{\theta_{\text{RL}}}$.

\subsection{Inference}

During inference, given a question $\mathcal{Q}$, a set of views $V=\{v_n\}_{n=1}^{N}$, and view transformation relationships $E$, our well-trained model $\pi_{\theta_{\mathrm{RL}}}$ executes the agentic spatial reasoning pipeline to generate the answer.
The model starts with an empty cognitive map $s_1=\emptyset$ and iteratively performs view retrieval and cognitive map updating until it stops exploring.
At each step, the model generates a retrieval action $a_{t+1} \sim \pi_{\theta_{\mathrm{RL}}}(s_t, \mathcal{Q}, E)$ to fetch a relevant view $v_i \in V$, then updates the cognitive map to $s_{t+1}$ by integrating the observation from $v_i$.
Finally, the model generates the answer $\hat{Y}$ based on the accumulated cognitive map $s_T$.

\section{Experimental Results}
\begin{table*}[!t]
\centering
\caption{
Accuracy comparison on the MindCube-Tiny benchmark.
The best results are highlighted in \textbf{bold} and the second best results are \underline{underlined}.
Some previous results are derived from MindCube~\cite{mindcube}.
}
\vspace{-2mm}
\small 
\begin{tblr}{
  colspec = {l c c c c c},
  row{1-Z}={rowsep=0.1em},
  column{1-Z} = {leftsep=0.4em, rightsep=0.4em},
  vline{2} = {1}{1-Z}{0.3pt}, 
  vline{2} = {2}{1-Z}{0.3pt}, 
  rows = {m}, 
  row{3,6,9,12} = {font=\itshape\color{graytext}},
  row{Z} = {bg=oursbg},
  row{4,7,10,13,15} = {bg=evengray},
  row{1,2} = {font=\bfseries, bg=headergray}, 
  hline{1,Z} = {1.5pt}, 
  hline{3,6,9,12} = {1}{1-Z}{0.3pt}, 
  hline{3,6,9,12} = {2}{1-Z}{0.3pt}, 
  cell{1}{1} = {r=2}{c}, 
  cell{1}{2} = {c=5}{c}, 
  cell{3,6,9,12}{1} = {c=6}{l}, 
}
    Methods & MindCube-Tiny Benchmark & & & & \\
     & Features & Overall $\uparrow$ & Rotation $\uparrow$ & Among $\uparrow$ & Around $\uparrow$ \\
    Baseline & & & & & \\
    Random (chance) & -- & 32.35  & 36.36 & 32.29 & 30.66 \\
    Random (frequency) & -- & 33.02  & 38.30 & 32.66 & 35.79 \\
    Open-Source VLMs & & & & & \\
    Qwen2.5-VL-7B-Instruct~\cite{qwen2.5-vl} & Passive        & 29.26  & 38.76 & 29.50 & 21.35  \\
    Qwen2.5-VL-3B-Instruct~\cite{qwen2.5-vl} & Passive        & 33.21  & 37.37 & 33.26 & 30.34  \\
    Proprietary Models & & & & \\
    GPT-4o~\cite{GPT-4o} & Passive &  38.81 & 32.65 & 40.17 & 29.16 \\
    Claude-4-Sonnet-20250514~\cite{claude35_sonnet} & Passive   &  44.75  & 48.42 & 44.21 & 47.62 \\
    Spatial VLMs & & & & \\
    Spatial-MLLM~\cite{SpatialMLLM} & Passive, 2D+3D & 32.06 & 38.39 & 20.92 & 32.82 \\
    Space-Qwen~\cite{SpatialVLM} & Passive & 33.28  & 38.02 & 33.71 & 26.32 \\
    MindCube$_{\textrm{Qwen2.5-VL-3B}}$~\cite{mindcube} & Passive, Cog. Map  & 70.7  & 48.0 & 79.2 & 68.4 \\
    3DThinker$_{\textrm{Qwen2.5-VL-3B}}$~\cite{3DThinker} & Passive, 3D Rec. & \underline{75.2} & \underline{55.5} & \textbf{81.8} & \underline{75.2} \\
    Ours & Active, Cog. Map & \textbf{80.5}\up{5.3} & \textbf{85.0}\up{29.5} & \underline{81.0}\down{0.8} & \textbf{75.6}\up{0.4} \\
\end{tblr}
\label{tab:main_table}
\end{table*}

In this section, we conduct extensive experiments to evaluate the effectiveness of our proposed method, including comparisons with state-of-the-art methods, ablation studies on the reward function, post-training stages, retrieval strategy, and memory mechanism.

\subsection{Setup}\label{sec:exp:setup}
\noindent\textbf{Dataset.}
Our experiments are conducted on the MindCube dataset~\cite{mindcube}. Specifically, the training set of MindCube contains 10,000 spatial reasoning questions.
Each question is associated with at most four cross-view images.
The testing set of MindCube, \ie, MindCube-Tiny, consists of 1,050 QA pairs~\cite{mindcube} and comprises multiple-choice questions with several options.
MindCube systematically challenges models across five key dimensions: camera movements, spatial patterns, ``what-if'' dynamic transformations, relational queries, and perspective-taking.
The benchmark therefore enables a comprehensive evaluation of a model's ability to understand complex and dynamic spatial scenes.
It is divided into three subsets, \ie, \textsc{Rotation}, \textsc{Among}, and \textsc{Around}, which correspond to different camera motions. In the MindCube-Tiny benchmark, 600 from the \textsc{Among}, 250 from \textsc{Around}, and 200 from \textsc{Rotation}. 

\noindent\textbf{Evaluation Metrics.}
We evaluate performance using question-answering accuracy. 
For a comprehensive assessment, we also report macro precision, recall, and F1-score.

\noindent\textbf{Compared Methods.} We compare our method with random baselines, including Random (chance) that uniformly selects from answer options and Random (frequency) that selects based on training label distribution, open-source VLMs~\cite{qwen2.5-vl}, property models~\cite{GPT-4o,claude35_sonnet}, and spatial VLMs~\cite{SpatialMLLM,SpatialVLM,mindcube,3DThinker}.
In particular, we primarily focus on comparing with state-of-the-art spatial VLMs, \ie, MindCube~\cite{mindcube} and 3DThinker~\cite{3DThinker}.
MindCube~\cite{mindcube} employs a passive perception and models a cognitive map to enhance spatial reasoning.
3DThinker\cite{3DThinker} also utilizes passive perception, and it leverages a 3D reconstruction auxiliary task to improve spatial understanding.

\noindent\textbf{Implementation Details.}
Our model is optimized using LLaMA-Factory~\cite{LlamaFactory} and veRL~\cite{verl}.
We use official Qwen2.5-VL-3B-Instruct~\cite{qwen2.5-vl} checkpoint as base model for post-training, following the same implementation in MindCube~\cite{mindcube} for fair comparison.
During training, we use 8 NVIDIA A100 GPUs with 80GB memory to optimize all parameters of the VLM.
For the SFT phase, we optimize the base model with a learning rate of $1\times 10^{-5}$.
The SFT stage is finished in 22 hours.
For the RL stage, the learning rate is set to $1\times 10^{-6}$.
We set the batch size to 128, and the number of rollouts $G$ in~\cref{eq:grpo_objective} to 8. 
We train the model for 1 day in the RL stage.
During inference, we finish the evaluation on 1,050 benchmark questions within 20 minutes.

\subsection{Main Results}

We compare our method with prior approaches on the MindCube-Tiny benchmark~\cite{mindcube}.
As summarized in \cref{tab:main_table}, our methodology achieves the highest accuracy overall and maintains consistent gains across the \textsc{Rotation}, \textsc{Among}, and \textsc{Around} subsets, indicating robust spatial reasoning under diverse camera movements.
Specifically, we achieve \normalup{5.3} gains overall against previous best approach, 3DThinker$_{\textrm{Qwen2.5-3B}}$~\cite{3DThinker}, and surpass MindCube~\cite{mindcube} by \normalup{9.8}.

This achievement originates from our advanced pipeline and training design.
MindCube and 3DThinker rely on a sparse reward signal, \ie, 0/1 reward that stands for correct or not, while we provide a dense reward during the RL stage.
Besides, we perform an additional RFT experiment on the \textsc{Rotation} as it presents a unique challenge that it lacks a central visual anchor and involves orthogonal views with minimal visual overlap, which fundamentally simulates embodied exploration.
Unlike object-centric settings (\textsc{Among} and \textsc{Around}), \textsc{Rotation} forces the agent to stitch fragmented egocentric views into a coherent environment.
This makes it a critical proxy for real-world robotic navigation, where agents must overcome limited field-of-view constraints.
We conduct RFT with passive (input all views) and active (perceive views iteratively) perception paradigms on the \textsc{Rotation} set and the results are reported in \cref{tab:ablation} (top).
The models are trained starting from the Qwen2.5-VL-3B-Instruct pretrained checkpoint without SFT.
Our method gets impressive \normalup{29.5} improvement in \cref{tab:main_table} and \normalup{11} accuracy increment in \cref{tab:ablation} (top), as well as gains in precision, recall, and F1-score.
The passive perception methods (MindCube and 3DThinker) struggle to align these disjointed images into a coherent 360\ensuremath{^\circ} representation when processing them simultaneously.
Conversely, our active paradigm transforms this into a sequential task.
By explicitly linking perceiving actions to visual feedback, the model establishes a robust egocentric reference frame.
This step-by-step grounding lowers the cognitive burden of spatial construction, enabling robust reasoning even without shared visual features across views.

\begin{figure}[t]
\centering
\includegraphics[width=.49\linewidth]{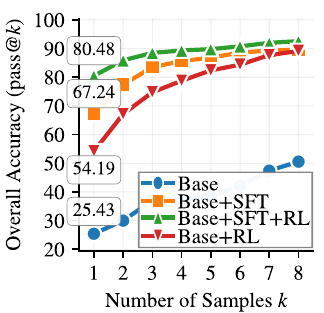}
\hfill
\includegraphics[width=.49\linewidth]{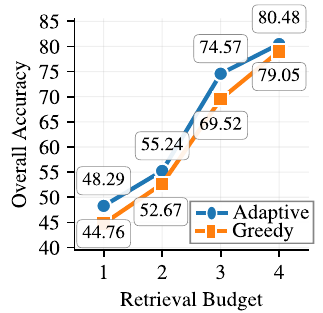}
\caption{
Pass@k accuracy curves demonstrating the contributions of SFT and RFT stages (left) and comparison of the adaptive and greedy retrieval strategies (right).
}
\label{fig:combined_curves}
\end{figure}

\begin{table}[t]
\centering
\caption{
Ablation studies on different components: perception pipeline (top), reward component ablation (middle), and memory mechanism (bottom).
}
\vspace{-2mm}
\small
\begin{tblr}{
  colspec = {l c c c c},
  row{1-Z}={rowsep=0.1em},
  column{1-Z} = {leftsep=0.4em, rightsep=0.4em},
  vline{2} = {1}{1-Z}{0.3pt},
  vline{2} = {2}{1-Z}{0.3pt},
  rows = {m},
  row{1} = {font=\bfseries, bg=headergray},
  hline{1,Z} = {1.5pt},
  hline{2,5,10} = {1}{1-Z}{0.3pt},
  hline{2,5,10} = {2}{1-Z}{0.3pt},
  cell{2,5,10}{1} = {c=5}{l},
  cell{1}{1} = {c=1}{c},
  row{2,5,10} = {font=\itshape\color{graytext}},
}
  Setting & Acc. $\uparrow$ & Pre. $\uparrow$ & Rec. $\uparrow$ & F1 $\uparrow$ \\
  Study of Perception Pipeline & & & & \\
  Passive       & 27.5 & 26.7 & 27.3 & 26.6 \\
  Active (Ours) & \textbf{38.5}\up{11} & \textbf{30.6}\up{3.9} & \textbf{29.2}\up{1.9} & \textbf{29.0}\up{2.4} \\

  Reward Component Ablation & & & & \\
  Ours Full  & \textbf{80.4} & \textbf{79.2} & \textbf{79.3} & \textbf{79.2} \\
  Ours w/o $R_{\textrm{retrieval}}$ & 72.5\down{7.9} & 57.4\down{21.8} & 57.0\down{22.3} & 57.1\down{22.1} \\
  Ours w/o $R_{\textrm{cogmap}}$    & 72.6\down{7.8} & 57.3\down{21.9} & 57.0\down{22.3} & 57.1\down{22.1} \\
  Ours w/o $R_{\textrm{SAC}}$ & 70.2\down{10.2}& 55.5\down{23.7} & 54.5\down{24.8} & 54.9\down{24.3} \\

  Study of Memory Mechanism & & & & \\
  Context & 50.9 & 36.2 & 34.8 & 34.8 \\
  Cog. Map (Ours) & \textbf{54.2}\up{3.3} & \textbf{40.6}\up{4.4} & \textbf{39.4}\up{4.6} & \textbf{39.5}\up{4.7} \\
\end{tblr}
\label{tab:ablation}
\end{table}

\noindent\textbf{Visualization.}
\begin{figure*}[!t]
\centering
\includegraphics[width=1.\linewidth]{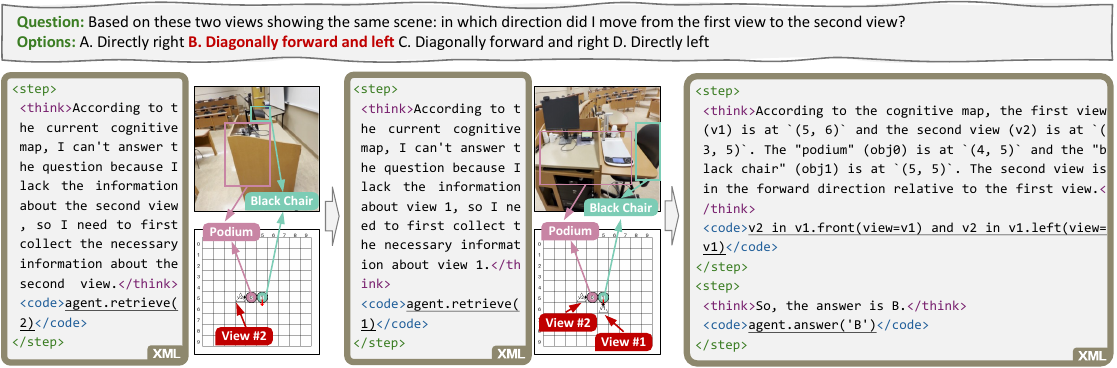}
\vspace{-2mm}
\caption{
A case study visualizing our method's active exploration process.
Our model retrieves two views and builds a dynamic cognitive map that parameterizes the layout of the scene.
During reasoning, the model outputs SAC that states \texttt{v2} is in front of \texttt{v1,} meanwhile to the left of \texttt{v1}.
Therefore, the correct answer ``B. Diagonally forward and left'' is inferred.
}
\label{fig:visualization}
\end{figure*}
We present a visualized case study in \cref{fig:visualization}.
Our model retrieves two views and constructs a cognitive map.
The cognitive map clearly illustrates the layout of the salient objects and cameras in the scene, \ie, \textit{black chair} and \textit{podium}.
Finally, the model reasons based on the cognitive map and correctly answers the question.
During reasoning the answer, our model generates code to describe the spatial relationships between objects, \ie, \texttt{v2 in v1.front(view=v1) and v2 in v1.left(view=v1)}.

\subsection{Ablation Studies}

We conduct comprehensive ablation experiments to isolate the contributions of individual components in our framework: reward, cognitive map, post-training stages, and retrieval strategy.

\noindent\textbf{Reward Component Ablation.}
We assess the effect of our reward function used in the RL stage.
As shown in \cref{tab:ablation} (middle), we present the overall accuracies obtained by ablating each component in \cref{eq:reward_function}.
Each component plays a critical role: ablating $R_{\textrm{SAC}}$ causes the most significant performance drop of \normaldown{10.2} in accuracy. The substantial degradation in macro precision, recall, and F1 scores (by approximately 22 points) when removing reward components reveals two critical issues.
First, without intermediate rewards, the cumulative error propagation severely impacts final reasoning accuracy.
This occurs because downstream decisions depend on upstream information gathering, making early errors propagate through the entire reasoning pipeline.
Second, the lack of syntactic constraints leads to action execution failures during the reasoning process.
These failures result in invalid outputs (\emph{e.g.}, ``None'' instead of valid answer choices), which dramatically reduce macro-level metrics.
The contrast between accuracy decline (7-10 points) and macro-metric collapse (22 points) highlights this output validity issue.

\noindent\textbf{Study of Memory Mechanism.}
As shown in \cref{tab:ablation} (bottom), replacing the naive input-output context (accumulating historical QA context as plain text) with our cognitive map yields an accuracy gain of \normalup{3.3}.
This improvement proves that simply accumulating historical context is insufficient for spatial reasoning due to viewpoint misalignment, while the top-down view cognitive map allows the VLM to retain precise object locations across changing views.

\noindent\textbf{Study of SFT and RFT.}
To validate the contribution of each training stage in our training pipeline, we evaluate the pass@$k$ performance across four model configurations, \ie, Base (pretrained Qwen2.5-VL-3B model), Base+RL, Base+SFT, and Base+SFT+RL.
According to~\cite{uppper}, the pass@$k$ performance is an upper bound of a model's capacity with a large $k$ while RL can narrow the gap between the pass@$1$ performance and the pass@$k$ upper bound~\cite{rlvr-narrow}.
The pass@$k$ curves of the 4 models are shown in the left panel of \cref{fig:combined_curves}, with $k$ progressively increasing from 1 to 8.
As illustrated in the figure, the base model exhibits limited zero-shot capability with a pass@1 accuracy of only 25.4 and a pass@8 accuracy of 50.6.
This indicates that the base model lacks sufficient capacity for agentic spatial reasoning and cannot reliably generate valid SAC for RFT even with multiple attempts, highlighting the necessity of alignment techniques. While both SFT and RL independently improve performance, their impacts differ significantly.
The application of SFT alone yields a substantial gain, propelling the pass@1 accuracy to 67.2, whereas applying RL directly to the base model results in a lower pass@1 accuracy of 54.2.
This disparity suggests that supervised data is crucial for establishing the initial instruction-following baseline that RL alone cannot fully reconstruct. Crucially, the combination of these methods delivers the best performance.
The full Base+SFT+RL configuration achieves a pass@1 accuracy of 80.5, representing a \normalup{13.3} absolute improvement over SFT alone.
Furthermore, analyzing the performance trend as $k$ increases reveals the model's sample efficiency.
Unlike the base model, which relies heavily on resampling to boost performance, the Base+SFT+RL model exhibits a flatter curve with early saturation.
Even at $k=8$, the gap between the base and the aligned models remains significant, confirming that our full pipeline not only maximizes peak accuracy but also minimizes the need for extensive test-time compute.

\noindent\textbf{Study of Retrieval Strategy.}
To examine how view retrieval impacts spatial reasoning, we compare two retrieval configurations:
(1) adaptively perceive views (ours),
(2) and greedily perceive unobserved views.
The right panel of \cref{fig:combined_curves} illustrates the overall accuracies achieved under different retrieval strategies and retrieval budgets.
As demonstrated in the figure, the accuracy of adaptive and greedy retrieving increases as the retrieval budget increases because more retrieval actions integrate more information to answer a question.
Comparing the two strategies, we can observe that our adaptive retrieval strategy always surpasses the greedy one.
It is important to note that the discrepancy is particularly significant with lower retrieval budgets (1, 2, and 3).
This indicates that the views perceived by our model are more relevant to the question to be answered.

\begin{figure}[!t]
\centering
\includegraphics[width=1.\linewidth]{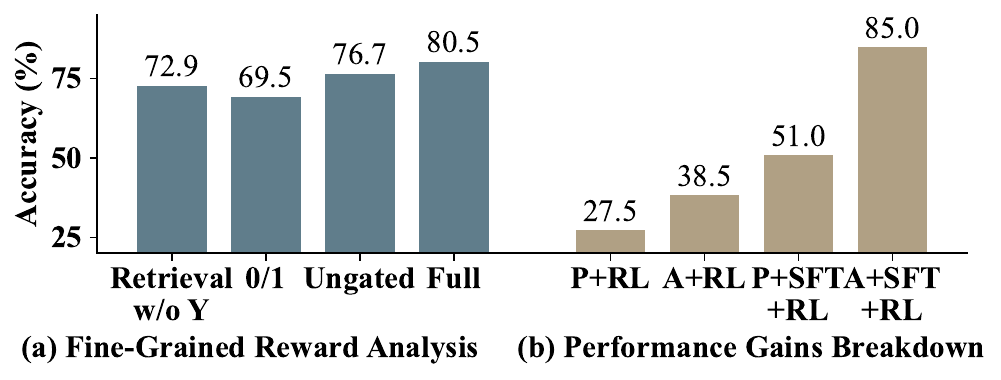}
\vspace{-2mm}
\caption{
Fine-grained reward analysis (left): removing $Y$ from retrieval supervision in $R_{\text{retrieval}}$ (Retrieval w/o Y), using only the outcome term $\mathbbm{1}_{\text{correct}}$ as reward (0/1), dropping the gating factor $\mathbbm{1}_{\text{correct}}$ in \cref{eq:reward_function} (Ungated), and our full reward (Full).
Performance gains breakdown (right): passive RL-only (P+RL), active RL-only (A+RL), passive with SFT followed by RL (P+SFT+RL), and the full active pipeline with SFT and RL (A+SFT+RL).
}
\label{fig:rebuttal_exp}
\end{figure} 

\noindent\textbf{Fine-Grained Reward Analysis.}
We further dissect two structural aspects of our reward: the reliance on privileged supervision from the dataset metadata except the final answer $Y$ during retrieval, and the role of the outcome-gating mechanism $\mathbbm{1}_{\text{correct}}$ in \cref{eq:reward_function}.
As shown in \cref{fig:rebuttal_exp} (left), removing $Y$ from the retrieval reward $R_{\text{retrieval}}$ (``Retrieval w/o Y'') causes a substantial drop of \normaldown{7.6} in accuracy, indicating that $Y$ serves as the primary supervisory signal while the remaining metadata plays only an auxiliary role. This suggests that our method does not heavily depend on dataset-specific metadata and can generalize to settings where such privileged information is unavailable.
Replacing the full reward with a pure 0/1 outcome reward (``0/1'') degrades performance by \normaldown{11.0}, confirming that dense intermediate feedback is essential beyond sparse outcome signals.
Removing the outcome gate (``Ungated'') results in a \normaldown{3.8} drop, suggesting that while intermediate rewards alone carry useful signal, the gating mechanism helps stabilize training by preventing reward hacking.

\noindent\textbf{Performance Gains Breakdown.}
To disentangle the contributions of active perception and supervised fine-tuning, we conduct a matched comparison using the same SFT and RL recipe for both passive and active perception paradigms on the challenging \textsc{Rotation} set.
As shown in \cref{fig:rebuttal_exp} (right), active perception consistently outperforms its passive counterpart: A+RL surpasses P+RL by \normalup{11.0} score, and A+SFT+RL surpasses P+SFT+RL by \normalup{34.0} score, confirming that agentic exploration provides complementary gains beyond what SFT alone can offer.
Meanwhile, SFT consistently improves both paradigms: P+SFT+RL exceeds P+RL by \normalup{23.5} points, and A+SFT+RL exceeds A+RL by \normalup{46.5} points, demonstrating that cold-start supervised data is essential for establishing the foundational capability.

\noindent\textbf{Robustness to Cognitive Map.}
We analyze the relationship between positional errors in the predicted cognitive map and final prediction correctness.
As shown in \cref{tab:cogmap_confusion}, among samples with positional errors, more than half still achieve correct predictions.
Conversely, even among correctly predicted samples, 15.7\% still exhibit positional errors in the cognitive map.
This indicates that the model does not rely strictly on precise coordinates and our cognitive map is robust in the whole pipeline.
We further compare the $10\times 10$ grid with a finer $20\times 20$ resolution in \cref{tab:cogmap_resolution}: increasing resolution yields 78.76\% accuracy, slightly lower than our 80.48\%, confirming that higher resolution adds visual complexity without providing additional useful information.

\begin{figure}[ht]
\centering
\begin{minipage}{0.48\linewidth}
\centering
\captionof{table}{Confusion matrix of positional errors \vs prediction.}
\label{tab:cogmap_confusion}
\small
\begin{tblr}{
  colspec = {c c c},
  row{1-Z}={rowsep=0.1em},
  column{1-Z} = {leftsep=0.5em, rightsep=0.5em},
  rows = {m},
  row{1} = {font=\bfseries, bg=headergray},
  hline{1,Z} = {1.5pt},
  vline{2} = {1}{1-Z}{0.3pt},
}
  {Positional \\ Error} & {Pred. \\ Correct} & {Pred. \\ Wrong} \\
  Yes & 133 & 114 \\
  No  & 712 & 91  \\
\end{tblr}

\end{minipage}
\hfill
\begin{minipage}{0.48\linewidth}
\centering
\captionof{table}{Comparison of different cognitive map resolutions.}
\label{tab:cogmap_resolution}
\small
\begin{tblr}{
  colspec = {c c},
  row{1-Z}={rowsep=0.1em},
  column{1-Z} = {leftsep=0.5em, rightsep=0.5em},
  rows = {m},
  row{1} = {font=\bfseries, bg=headergray},
  hline{1,Z} = {1.5pt},
}
  {Grid \\ Resolution} & Acc. (\%) \\
  20 & 78.76 \\
  10 (Ours) & \textbf{80.48} \\
\end{tblr}

\end{minipage}
\end{figure}

\begin{table}[t]
\centering
\caption{Performance comparison on the EmbodiedBench benchmark~\cite{EmbodiedBench}.
We evaluate Qwen2.5-VL-3B-Instruct as the base model and our model trained on MindCube (Base+SFT+RL) without any finetuning on EmbodiedBench.
}
\small 
\begin{tblr}{
  colspec = {l c c c c c c},
  row{1-Z}={rowsep=0.1em},
  column{1-Z} = {leftsep=0.1em, rightsep=0.1em},
  vline{2} = {1}{1-Z}{0.3pt}, 
  vline{2} = {2}{1-Z}{0.3pt}, 
  rows = {m}, 
  row{1} = {font=\bfseries, bg=headergray}, 
  row{3} = {bg=evengray},
  hline{1,Z} = {1.5pt}, 
}
  Model & Avg. $\uparrow$ & Base $\uparrow$ & Com. $\uparrow$ & Comp. $\uparrow$ & Vis. $\uparrow$ & Long $\uparrow$ \\
  Base & 6.3 & 5.0 & 11.7 & \textbf{8.3} & 5.0 & 1.7 \\
  Ours & \textbf{6.7} & \textbf{15.0} & 0.0 & 0.0 & \textbf{16.7} & 1.7 \\
  LM+Ours & -- & -- & \textbf{15.0} & 6.7 & -- & --
\end{tblr}
\label{tab:embodied_bench}
\end{table}
\section{Discussion}

\noindent\textbf{Generalizability to the Embodied Navigation Task.}
To assess the generalization capability to real-world embodied navigation, we further conduct experiments on the EmbodiedBench benchmark~\cite{EmbodiedBench}. 
The agents must locate target objects and navigate to their vicinity by executing low-level physical actions in a simulated environment.
It focuses on executing actions to complete tasks in environments, while our spatial QA task emphasizes enhancing models' spatial cognition capabilities. We compare our model trained on MindCube with the baseline Qwen2.5-VL-3B-Instruct pretrained model.
None of the models is finetuned on EmbodiedBench.
As shown in \cref{tab:embodied_bench}, our model achieves substantial improvements in the \textsc{Base} and \textsc{Visual} settings, but drops to zero on \textsc{Common} and \textsc{Complex}.
This gap stems from a task mismatch: EmbodiedBench primarily tests language-grounded instruction understanding (\eg, commonsense and complex linguistic reasoning), whereas our model is specialized for spatial reasoning.
A practical remedy is to combine our model with a language model ahead for instruction comprehension.
As shown in \cref{tab:embodied_bench} (LM+Ours), this simple combination raises \textsc{Common} from 0.0 to 15.0 and \textsc{Complex} from 0.0 to 6.7, confirming that the bottleneck lies in language understanding rather than spatial reasoning.

Our future work may extend this framework to more complex embodied tasks, such as object manipulation, embodied navigation, multiagent collaboration, and long-horizon planning in dynamic environments.

\section{Conclusion}

In this work, we presented an agentic spatial reasoning framework that enables VLMs to actively explore 3D scenes through selective view retrieval and mental spatial reasoning.
We formulated spatial reasoning as a sequential decision-making problem, where the model maintains a continually updated cognitive map and selectively queries relevant views based on task demands.
To guide the learning process, we introduced spatial assertion codes (SAC) collaborating with the cognitive map to evaluate intermediate reasoning steps.
Through synthetic data generation for supervised fine-tuning and reinforcement finetuning, we achieved state-of-the-art performance on the MindCube benchmark.

\nocite{DC-SAM,RDCLR,HP-MCoRe,treesearchgen,T2SG,SGFormer}

\section*{Acknowledgements}
This work is partly supported by the Funds for the National Natural Science Foundation of China under Grant 62572072,  and Beijing Natural Science Foundation (L243027).

\section*{Impact Statement}
This paper presents work whose goal is to advance the field of Machine Learning.
There are many potential societal consequences of our work, none which we feel must be specifically highlighted here.

\bibliography{main}
\bibliographystyle{icml2026}

\end{document}